\documentclass[final]{l4dc2025}
\usepackage{booktabs}
\usepackage{multirow}
\usepackage[super]{nth}
\usepackage[amssymb]{SIunits}
\usepackage{verbatim}
\usepackage{bm}
\usepackage{booktabs}
\usepackage[capitalise]{cleveref}
\usepackage{todonotes}
\usepackage{pifont}
\usepackage[normalem]{ulem}
\usepackage{xcolor}

\title[Imperative MPC]{Imperative MPC: An End-to-End Self-Supervised Learning with Differentiable MPC for UAV Attitude Control}
\usepackage{times}

\author{%
 \Name{Haonan He} \Email{haonanh@alumni.cmu.edu}\\
 \Name{Yuheng Qiu} \Email{yuhengq@andrew.cmu.edu}\\
 \addr Department of Mechanical Engineering, Carnegie Mellon University
 \AND
 \Name{Junyi Geng} \Email{jgeng@psu.edu}\\
 \addr Department of Aerospace Engineering, Pennsylvania State University%
}

\begin{document}

\maketitle 


\begin{abstract}%
Modeling and control of nonlinear dynamics are critical in robotics, especially in scenarios with unpredictable external influences and complex dynamics. Traditional cascaded modular control pipelines often yield suboptimal performance due to conservative assumptions and tedious parameter tuning. Pure data-driven approaches promise robust performance but suffer from low sample efficiency, sim-to-real gaps, and reliance on extensive datasets. Hybrid methods combining learning-based and traditional model-based control in an end-to-end manner offer a promising alternative.
This work presents a self-supervised learning framework combining learning-based inertial odometry (IO) module and differentiable model predictive control (d-MPC) for Unmanned Aerial Vehicle (UAV) attitude control. The IO denoises raw IMU measurements and predicts UAV attitudes, which are then optimized by MPC for control actions in a bi-level optimization (BLO) setup, where the inner MPC optimizes control actions and the upper level minimizes discrepancy between real-world and predicted performance. The framework is thus end-to-end and can be trained in a self-supervised manner. This approach combines the strength of learning-based perception with the interpretable model-based control. Results show the effectiveness even under strong wind. It can simultaneously enhance both the MPC parameter learning and IMU prediction performance.
\end{abstract}


\section{Introduction} \label{sec:intro}
Modeling and control of nonlinear dynamics precisely is essential for robotics, particularly in challenging settings like aerial robotics \citep{mousaei2022design}, aerial manipulation tasks \citep{geng2020cooperative}. Such environments often feature complex nonlinear effects like coupled translational and rotational motion, or unpredictable external influences, complicating control design. 
Conventional methods like state feedback \citep{mellinger2011minimum, blondin2019control, kozak2016pid} or optimization-based control \citep{garcia1989model, ji2020robust} require full knowledge of the system model. However, it is challenging to fully capture unpredictable factors such as wind gusts, boundary layer effects, uneven terrain, or hidden dynamics of chaotic nonlinear effects. These methods end up facing significant challenge and requiring tedious parameters tuning \citep{borase2021review, li2016research, alhajeri2020tuning}. 

Traditional control pipelines are typically modular, where perception \& state estimation, planning and control are designed separately and then integrated, which often leads to suboptimal performance, as conservative assumptions in one component may cause local minima. The overall performance is also constrained by the limitation of individual modules. In addition, iterative design update is required, where parameters of one module are fine-tuned while others are frozen, making the process tedious and time-consuming \citep{liu2023path, hanover2024autonomous}.
In contrast, data-driven approaches have gained more interest, which directly map sensor observation to action and offer robust learning-based perception. However, these methods usually require large amount of data, have low sample efficiency, and pose significant sim-to-real gap \citep{rawles2024androidinthewild, ma2024hierarchical}. For example, some researchers train control policies by imitating expert actions. However, this approach demands extensive datasets to generalize across diverse environments \citep{xing2024bootstrapping}. Another route is reinforcement learning (RL) and its variants \citep{williams2017information}, though promising, often exhibit slow convergence, low sample efficiency, and overfitting risks. Additionally, reward shaping can inadvertently introduce biases and result in suboptimal policies.


Recently, researchers have explored hybrid approaches that integrate learning-based methods with model-based approaches by embedding differentiable modules into learning frameworks. While much of this work focuses on perception, state estimation, and path planning, fewer studies address control. \cite{amos2018differentiable} developed a differentiable model predictive control (MPC), combining physics-based modeling with data-driven methods to enable end-to-end learning of dynamics and control policies. However, many prior approaches rely on expert demonstrations or labeled data for supervised learning \citep{jin2020pontryagin, song2022policy}, which, while effective on training datasets, struggle to generalize to unseen environments or handle external disturbances.

To address these issues, we propose a self-supervised learning framework combining learning-based perception with traditional model-based control. MPC can optimize control actions over a time horizon while satisfying system constraints. However, applying MPC in highly dynamic environments like wind gusts or moving obstacles in unmanned aerial vehicle (UAV) control remains challenging to balance optimality with real-time reactivity.
For perception, UAVs face size, weight, power, and cost (SWAP-C) constraints \citep{mohsan2022towards, youn2021collision}, making inertial odometry (IO) based on lightweight, low-cost inertial measurement units (IMUs) ideal. Unlike vision or LiDAR sensors, IMUs avoid visual degradation factors such as motion blur or dynamic object interference \citep{zheng2007computer, zhou2023deblurring}, which are common in agile flights. However, traditional IO suffers from sensor noise and drift, leading to suboptimal performance \citep{tedaldi2014robust}. Learning-based IO, alternatively, has gained attention for its ability to model and correct inherent data imperfections, thereby improving reliability and accuracy \citep{qiu2023airimu}.
We incorporate differentiable MPC (d-MPC) into a self-supervised learning framework based on our prior work imperative learning (IL) \citep{wang2024imperative}, which we term imperative MPC. In our approach, a learning-based IO denoises raw IMU measurements and predicts UAV attitudes, then used by MPC for attitude control in a bi-level optimization (BLO) setup. The inner level solves standard MPC and the upper level optimizes discrepancy between control performance of the real vehicle and the predicted model. The whole framework is trained in a self-supervised manner. This joint training improves both MPC parameters and IMU prediction performance, leveraging robust learning-based perception and interpretable model-based control without modular separation.

\subsection{Related Works} \label{subsec:related}
\textbf{Learning-based Control}. Hybrid approaches integrating learning-based methods with traditional model-based control become an emerging trend. These approaches range from combining learning residual dynamics with nominal dynamics and adaptive control to address real-time disturbances and uncertainties \citep{o2022neural}, or incorporate large, complex neural network architectures as surrogate dynamics within an MPC \citep{salzmann2023real}, to policy search to optimize MPC parameters for improved adaptability \citep{song2022policy}. However, these methods often follow a cascaded design, requiring separate stages for model learning and controller design.
Another direction leverages physics-informed neural networks, embedding traditional control principles into loss functions. For instance, \cite{mittal2020neural} integrates Lyapunov functions into the MPC framework, using neural networks to learn globally safe control policies. However, pure data-driven model introduces inaccuracies in learned Lyapunov functions or system dynamics, which degrade performance in dynamic environments.
Differentiable optimization has recently gained attention for its end-to-end nature. \cite{jin2020pontryagin} proposed Pontryagin Differentiable Programming (PDP), embedding system dynamics and control strategies into a differentiable framework using Pontryagin’s Minimum Principle (PMP). However, it is limited to local minima due to problem nonconvexity and first-order approximations. Furthermore, all these approaches heavily rely on supervised learning with substantial data and focus primarily on control, often neglecting upstream perception or planning in the autonomy pipeline.

\textbf{Hybrid Learning and Optimization Framework for Autonomy}.
Many studies explore hybrid approaches in autonomy pipelines. Some integrate optimization-based planners into end-to-end learning \citep{han2024learning, yang2023iplanner}, but often ignore dynamics constraints or rely on supervised learning, or focusing solely on planning without control. Others combine model-based control with learning for decision-making or control-action initialization \citep{baek2022hybrid, celestini2024transformer}, yet depend on modular training and offline-labeled data, limiting adaptability. Recent self-supervised methods, like \cite{li2024faster}'s offline-to-online RL, improve flexibility but retain offline constraints. Similarly, RL-MPC hybrids (e.g., \cite{romero2024actor}) leverage MPC's predictability and RL's exploration but assume known dynamics and face reward-shaping challenges. Our approach enables self-supervised reciprocal correction between neural and optimization components.

\vspace{-0.3mm}
\section{Preliminary} \label{sec:pre}

\subsection{Differentiable Model Predictive Control} \label{subsec:mpc}
In model based control, MPC has dominated across many industrial applications due to its robust ability of handling complex tasks. It leverages a predictive model of the dynamical system $F(\cdot)$ and repeatedly solves an online optimization problem to minimize objective $J$ in a receding horizon fashion with time horizon $N$, and produce the optimal states $\bm{x}$ and control $\bm{u}$ sequences ${\bm{\mu}}^* = \{{\bm{\mu}}_k\}_{1:N} = \{\bm{x}_k, \bm{u}_k\}_{1:N}$ (with only the first action is executed). Formally, at each time step, MPC solves an optimal control problem (OCP):
{\small
\begin{equation}
\begin{aligned}
    &\underset{\bm{x}_{1:N}, \bm{u}_{1:N}} {\text{min}} J = \sum_{k} c_k(\bm{x}_k, \bm{u}_k) \\
    &\textrm{s.t.} \quad  \bm{x}_{k+1} = F(\bm{x}_k, \bm{u}_k), \quad k = 1,\cdots,N \\
    &\quad \quad~\bm{x}_1 = \bm{x}_{init}; \; \bm{x}_k \in \mathcal{X}; \; \bm{u}_k \in \mathcal{U}
\end{aligned}
\label{eq: mpc01}
\end{equation}
}
where $\bm{x}_k$ and $\bm{u}_k$ are the state and control input at time step $k$, $\mathcal{X}$ and $\mathcal{U}$ denote constraints on valid states and controls.
From a learning perspective, the objective and dynamics as well as constraints of MPC may contains some unknown parameters, which can be predicted by some other autonomy components, e.g. $c_{\theta,k}$(\cdot), $F_{\theta}(\cdot)$, $\bm{x}_{init, \theta}$, with the learnable parameters $\theta$. The MPC thus can be integrated into a larger end-to-end systems. The learning process involves finding the derivatives of some loss function $l$, which are then used to update $\theta$. 
D-MPC allows gradient back-propagation through the traditional optimization process, enabling the end-to-end gradient can be obtained by:
\vspace{-1.5mm}
{\small
\begin{equation}
\begin{aligned}
    \nabla_{\theta} l = \nabla_{\theta} \bm{\mu} \, \nabla_{\bm{\mu}} l
\end{aligned}
\label{eq: mpc03}
\vspace{-1.5mm}
\end{equation}
}

The gradient consists of $\nabla_{\theta} \bm{\mu}$ and $\nabla_{\bm{\mu}} l$. $\nabla_{\bm{\mu}} l$ can typically be computed analytically, while calculating $\nabla_{\theta} \bm{\mu}$ is more challenging. It can be obtained from simply unrolling and maintaining the entire computational graph throughout the iteration process, which incurs significant challenges in terms of memory usage. It may also face issues related to gradient divergence or vanishment \citep{finn2017model}. Another approach is to use implicit function differentiation, leveraging the necessary conditions for optimality, allowing the gradient to be computed without unrolling \citep{dontchev2009implicit}. Denote $\xi \leq 0$ as the general constraints including the equality and inequality parts. The Lagrangian problem (\ref{eq: mpc01}) with $\lambda$ as the Lagrange multipliers can be formulated as:
\vspace{-1.8mm}
{\small
\begin{equation}
\begin{aligned}
    \mathcal{L}(\bm{\mu}, \lambda) = J(\bm{\mu}) + \lambda^\top \xi(\bm{\mu}, \theta)
\end{aligned}
\label{eq: mpc04}
\vspace{-1.5mm}
\end{equation}
}

Then we can get the corresponding KKT (Karush-Kuhn-Tucker) conditions:
\vspace{-2mm}
{\small
\begin{equation}
\begin{aligned}
    \nabla_{\bm{\mu}} J + \nabla_{\bm{\mu}} \xi \lambda^* &= 0, \quad D(\lambda^*) \xi (\bm{\mu}^*, \theta) = 0, \\
    \xi (\bm{\mu}^*, \theta) &\leq 0, \quad \lambda^* \geq 0.
\end{aligned}
\label{eq: mpc05}
\vspace{-2mm}
\end{equation}
}
where $D(\cdot)$ is a diagonal matrix derived from a vector. \cref{eq: mpc05} can be easily reformulated as:

\vspace{-2mm}
{\small
\begin{equation}
\begin{aligned}
    \begin{bmatrix}
    \mathrm{d} \bm{\mu} \\
    \mathrm{d} \lambda
    \end{bmatrix}
    = - \begin{bmatrix}
    \nabla_{\bm{\mu},\bm{\mu}} J + (\nabla_{\bm{\mu},\bm{\mu}} \xi \lambda^*)^\top & \nabla_{\bm{\mu}} \xi \\
    D(\lambda^*) \nabla_{\bm{\mu}} \xi & D(\xi)
    \end{bmatrix}^{-1}
    \begin{bmatrix}
    (\nabla_{\bm{\mu},\theta} \xi \lambda^*)^\top \\
    D(\lambda^*) \nabla_\theta \xi
    \end{bmatrix}
    \mathrm{d} \theta
\end{aligned}
\label{eq: mpc07}
\vspace{-0.5mm}
\end{equation}
}

We thus analytically derives $\nabla_{\theta} \bm{\mu}$, which subsequently enables us to compute the gradient $ \nabla_{\theta} L $ from \cref{eq: mpc03}. This enables efficient parameter updates within an end-to-end learning framework.

In this paper, we use the d-MPC developed in our previous work PyPose, an open-source library for robot learning \citep{wang2023pypose}. This module formulated the MPC as an iterative Linear Quadratic Regulator (iLQR) problem to tackle the non-convex and nonlinear scenarios. Unlike the existing method of KKT conditions, which is problem-specific, PyPose
implements one extra optimization iteration of iLQR at the stationary point in the forward pass, allowing automatic gradient calculation in the backward pass, which is more versatile and problem-agnostic with the price of small approximation error \citep{bolte2023one}.
This approach eliminates the need to compute gradients for the entire unrolled chain or derive analytical solutions tailored to each specific problem.

\subsection{Imperative Learning} \label{subsec:il}

Our prior work introduced Imperative Learning (IL), a framework that integrates a neural system, a reasoning engine, and a memory module to enhance robotic learning and decision-making \citep{wang2024imperative}. IL is formulated as a BLO problem, where the upper-level (UL) objective $ U $ optimizes neural parameters related to perception, while the lower-level (LL) objective $ L $ optimizes parameters related to reasoning and memory: 
\vspace{-1.5mm}
{\small
\begin{equation}
\begin{aligned}
    \min_{ \bm \psi  \doteq  [{\bm{\theta}}^\top,~{\bm{\gamma}}^\top]^\top} & U\left(f_{\theta}(\bm{z}), g({\bm{\mu}}^*), M({\bm{\gamma}}, {\bm{\nu}}^*)\right) \\
    \textrm{s.t.} \quad & \bm \phi^* \in \underset{\bm \phi \doteq  [{\bm{\mu}}^\top,~{\bm{\nu}}^\top]^\top}{\text{arg min}} L(f_{\theta}(\bm{z}), g({\bm{\mu}}), M({\bm{\gamma}}, {\bm{\nu}})) \\
    &\textrm{s.t.} \quad  \xi(M({\bm{\gamma}}, {\bm{\nu}}), {\bm{\mu}}, f_{\theta}(\bm{z})) = \text{ or }\leq 0
\end{aligned}
\label{eq: il01}
\vspace{-1.5mm}
\end{equation}
}
where $f_\theta(\bm{z})$ is the neural outputs such as semantic attributes, $\bm{z}$ represents the sensor measurements, and ${\bm{\theta}}$ is the perception-related learnable parameters. The reasoning engine is $g(f, M, {\bm{\mu}})$ with parameters ${\bm{\mu}}$, while the memory system is denoted as $M({\bm{\gamma}}, {\bm{\nu}})$ with perception-related parameters ${\bm{\gamma}}$ as well as reasoning-related parameters ${\bm{\nu}}$ \citep{wang2021unsupervised}. $\xi$ is a general constraint; and $\bm \psi \doteq [{\bm{\theta}}^\top, {\bm{\gamma}}^\top]^\top$ are stacked UL variables and $\bm \phi \doteq [{\bm{\mu}}^\top, {\bm{\nu}}^\top]^\top$ are stacked LL variables, respectively.
IL also leverages implicit differentiation to compute gradients for interdependent neural and symbolic parameters efficiently. The solution to IL \cref{eq: il01} mainly involves solving $\bm{\theta}$, $\bm{\gamma}$, $\bm{\mu}$, and $\bm{\nu}$.



The perception module extracts semantic features from sensor data, the symbolic reasoning module enforces logical or physical constraints, and the memory module stores relevant knowledge for long- and short-term retrieval when necessary such as in the application of SLAM. Through reciprocal learning, IL enables a self-supervised mechanism in which perception, reasoning, and memory modules co-evolve by iteratively refining each other to satisfy a global objective. This structure allows IL to combine the strength of both data-driven learning and physics knowledge. While the previous IL works mainly focus on SLAM, planning, and preliminary control \citep{yang2023iplanner, fu2024islam, wang2024imperative}, this work thoroughly introduces the usage of IL on control and provides high-fidelity validation results.

\section{Methodology} \label{sec:method}
\subsection{Pipeline Overview} \label{subsec:pipeline}
\begin{figure*}
    \centering
    \includegraphics[width=0.98\textwidth]{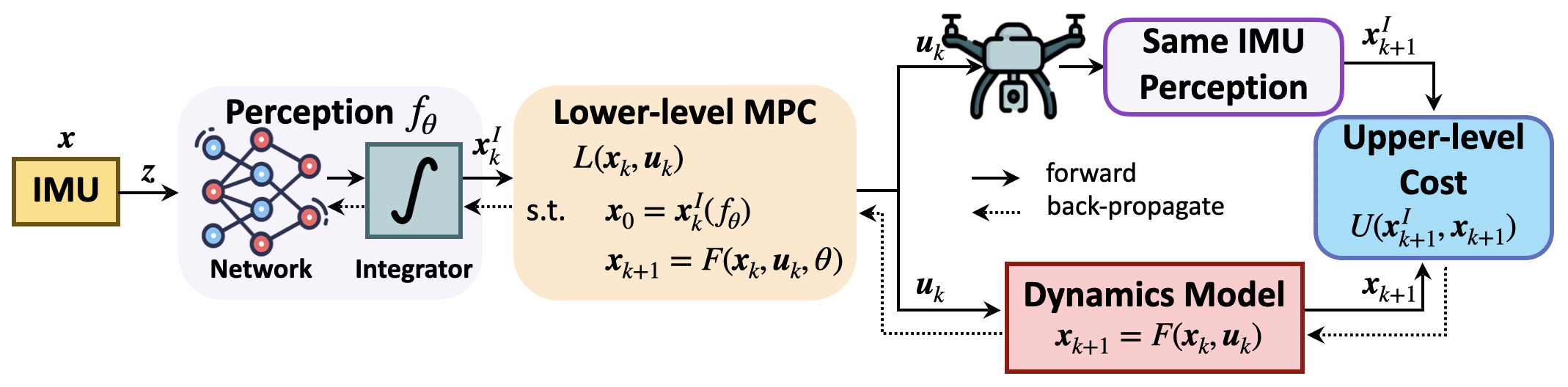} 
    \caption{The proposed framework. The IMU model predicts the current state $\bm{x}_k^I$. The d-MPC solves for the optimal action $\bm{u}_k$ under lower-level $L$, which controls the dynamics model to the next state ($\bm{x}_{k+1}$) and actuates the real system to next state measured by the same IMU ($\bm{x}^I_{k+1}$). The upper-level $U$ minimizes the discrepancy between $\bm{x}_{k+1}$ and $\bm{x}^I_{k+1}$.} 
    \label{fig:pipeline}
    \vspace{-8mm}
\end{figure*}

As illustrated in \cref{fig:pipeline}, our framework consists of a perception network $f_{\theta}(\bm{z})$, in this work is the learning-based IO,  and a physics-based optimization based on MPC. Specifically, the encoder $f(\cdot)$ takes into raw sensor measurements $\bm{z}$, and here are IMU measurements to predict the high-level parameters for MPC, such as objective function, dynamic model, initial condition or constraints. 
Here is UAV's current attitude as initial state $\bm{x}_0$.
Then, MPC takes the current attitude as initial states and solves for the the optimal states $\bm{x}$ and control $\bm{u}$ sequences ${\bm{\mu}}^* = \{{\bm{\mu}}_k\}_{1:N} = \{\bm{x}_k, \bm{u}_k\}_{1:N}$ over a time horizon $N$ under constraints of system dynamics $F(\cdot)$, actuator limit, and other factors. Then, only the first action is executed. The problem is be formulated as a BLO problem:
\vspace{-2mm}
{\small
\begin{equation}
\begin{aligned}
    \min_{{\bm{\theta}}} \quad & U\left(f_{\theta}(\bm{z}), g({\bm{\mu}}^*)\right) \\
    \textrm{s.t.} \quad & {\bm{\mu}}^* = \arg\min_{{\bm{\mu}}} L(f, g({\bm{\mu}})) \\
    &\textrm{s.t.} \quad  \bm{x}_{k+1} = F(\bm{x}_k, \bm{u}_k), \quad k = 1,\cdots,N \\
    &\quad \quad~\bm{x}_1 = \bm{x}(t_1); \; \bm{x}_k \in \mathcal{X}; \; \bm{u}_k \in \mathcal{U}
\end{aligned}
\label{eq:impc01}
\vspace{-2mm}
\end{equation}
}
where the lower level (LL) optimization is a standard MPC to output optimal control sequences $\bm{u}$ sequences ${\bm{\mu}}^*=\{{\bm{\mu}}_k\}_{1:N}$ with $\bm{x}_k$ and $\bm{u}_k$ are the state and control input at time step $k$, $\mathcal{X}$ and $\mathcal{U}$ are the constraints on valid states and controls. Notice in this scenario, there is no need for the memory system $M({\bm{\gamma}}, {\bm{\nu}})$ as in \cref{eq: il01}. The objective function $L$ is designed to minimize both tracking error and control effort: 
\vspace{-3.5mm}
{\small
\begin{equation}
    L({\bm{\mu}}_k) \doteq \sum_{k=0}^{N-1}\left(\Delta \bm \mu_k^\top\bm{Q}_k\Delta \bm \mu_k + \bm{p}_k \Delta \bm \mu_k \right)
\label{eq:imoc02}
\vspace{-2mm}
\end{equation}
}
Here $ \Delta \bm \mu_k = {\bm{\mu}}_k - {\bm{\mu}}_k^{\text{ref}}$, where ${\bm{\mu}}_k^{\text{ref}}$ represents the reference state trajectory, and $\bm{Q}_k$ and $\bm{p}_k$ are the weight matrices to balance tracking performance and control effort, respectively. The optimal control action $\bm{u}_k$ is then applied to both the modeled dynamics and the physical robot, resulting in the subsequent states $\bm{x}_{k+1}$ and $\bm{x}^I_{k+1}$ (the latter measured by the IMU network). 

The upper level (UL) cost is defined as the Euclidean distance between $\bm{x}_{k+1}$ and $\bm{x}^I_{k+1}$:
\vspace{-2mm}
{\small
\begin{equation}
    U({\bm{\theta}}) \doteq \| \bm{x}_{k+1}^I-\bm{x}_{k+1} \|_2.
\label{eq:impc03}
\vspace{-2mm}
\end{equation}
}

This discrepancy between the predicted states and the IMU network measurement captures the imperfectness of perception and dynamics model, suggesting the necessitates to adjust the learnable parameters ${\bm{\theta}}$ within both the network and MPC module. Notice that both \( x_k^{I} \) and \( x_{k+1}^{I} \) are indeed outputs of the perception module \( f_\theta \). However, gradients are explicitly backpropagated only through \( x_{k+1}^{I} \), treating the current-step state \( x_k^{I} \) as a fixed initial condition within the optimization loop. 

A critical step in (\ref{eq:impc01}) involves calculating the implicit gradient $\nabla_{\theta} \bm{\mu}^*$. We leveraging PyPose's d-MPC to eliminate the necessity to calculate gradients for the whole unrolled chain or analytical derivatives. The parameters in both IMU network and d-MPC can thus be jointly learned. Additionally, our framework reduces dependency on annotated labels by utilizing self-supervision from the physical model and achieves dynamically feasible predictions through the MPC module.


\subsection{Perception Networks--Learning based IO} \label{subsec:io}

For the perception model, we leverage a neural network to model the noise inhere in both the accelerometer and the gyroscope.
After filtering the noise, we employ the differentiable IMU pre-integrator from \cite{qiu2023airimu} to estimate the state $\bm{x}^I_{k+1} = f_\theta(z)$, where $z$ is the measurements of the IMU including accelerometer and the gyroscope.
The differentiability of the IMU pre-integrator enables gradient flow from the integrated state $\bm{x}^I_{k+1}$ back to the noise modeling network during training. 
To reduce computational complexity, we adopt a lightweight design: a 3-layer multi-layer perceptron (MLP) encoder extracts feature embeddings, followed by a 2-layer MLP decoder that predicts sensor noise. We randomly initialize the network weights and train the model from scratch.


\subsection{UAV Dynamics Model} \label{subsec:uav}

We consider a quadrotor UAV. The dynamics can be modeled using the Netwon-Euler method:
\vspace{-2mm}
{\small
\begin{align}\label{eq:dyn}
    m\Ddot{\mathbf{r}} = -mg\mathbf{z}_W + T\mathbf{z}_B \\
    \mathbf{J}\dot{\boldsymbol{\omega}} = - \boldsymbol{\omega}\times\mathbf{J}\boldsymbol{\omega} + \boldsymbol{\tau}\nonumber
\vspace{-2.3mm}
\end{align}
}
where $m$ is the UAV mass, $\mathbf{r} = [x, y, z]$ is the UAV position vector in world frame $\mathcal{W}$ with the $x_W-y_W-z_W$ axes, $g$ is the gravity vector, $T$ is the body thrust force magnitude, $\mathbf{J}$ is the moment of inertia matrix referenced to the center of mass along the body $x_B-y_B-z_B$ axes. $\boldsymbol{\omega}$ is the robot angular velocity of frame $\mathcal{B}$ in frame $\mathcal{W}$, $\boldsymbol{\tau} = [\tau_x, \tau_y, \tau_z]$ is the body moment. We also use 3-2-1 Euler angles to define the roll, pitch, and yaw angles $\boldsymbol{\Omega} = [\phi, \theta, \psi]$ as the orientation.
The state of the system is given by the position and velocity of the center of mass and the orientation and the angular velocity: $\mathbf{x} = [\mathbf{r}, \boldsymbol{\Omega}, \dot{\mathbf{r}}, \boldsymbol{\omega}]^\top$. The control input is defined as $\mathbf{u} = [T; \boldsymbol{\tau}]^\top$.

Eq. (\ref{eq:dyn}) will be incorporated as constraints $F(\cdot)$ of \cref{eq: mpc05} in (\ref{eq:impc01}). The optimization problem will be solved to obtain states and control inputs is dynamic feasible. The optimal control action $\bm{u}_k$ will be sent to both the modeled UAV dynamics and the real vehicle accordingly.

\section{Experiments} \label{sec:exp}

We conducted experiments to validate the effectiveness of iMPC for UAV attitude control. We evaluate the system performance under different initial condition, wind gusts, and model uncertainty.

\subsection{Experiment Setup} \label{subsec:setup}
We test our approach through both a customized Python simulation and a high-fidelity Gazebo simulator with widely used ROS and commercial PX4 Autopilot. The customized simulation is used to train our proposed framework and evaluate the controller performance. Validation in the Gazebo PX4 SITL (Software-in-the-Loop) simulation captures real-world effects, including sensor and actuator dynamics, environmental conditions, and communication delays, serving as a digital twin and final checkpoint prior to real-world flight testing.

The Python simulation consists of standard 6-degree-of-freedom (DoF) quadrotor UAV dynamics described in \cref{subsec:uav}, running at 1 K$\hertz$, and a simulated IMU sensor at 200 \hertz. To emulate real-world effects such as environmental disturbances, actuator uncertainties, sensor noise, and other unknown behaviors, we inject zero-mean Gaussian noise with a standard deviation of $1 \times 10^{-4}$ into the control input, and zero-mean Gaussian noise with a standard deviation of $8.73 \times 10^{-2}$ into the UAV attitude at 1 K$\hertz$. Additionally, we employ sensor noise models documented by the \cite{epsonG365IMU} IMU, including initial bias, bias instability, and random walk. This produces the same noise level as typical UAV tracking systems employing visual-inertial odometry \citep{qin2018vins} for attitude estimation. For wind disturbance experiments, we introduce external forces and torques based on the drag equation: $F = \frac{1}{2} \cdot C_d \cdot \rho \cdot A \cdot V^2$ and $\tau = r \times F$, where $C_d$ is the drag coefficient, $\rho$ air density, $A$ the frontal area of the object facing the wind, $V$ wind speed, and $r$ the lever arm length.

The Gazebo PX4 SITL simulation environment leverages the PX4 autopilot firmware as the low-level control and communicates with high-level decision-making via Mavros. Here, the MPC generates total thrust and angular rate commands, which are sent to the low-level controller. A 3DR iris quadrotor UAV model is used as the active agent, including all actuators and sensor plugins, such as the servo motors, IMU, etc.
We directly use the Gazebo wind gusts to inject wind disturbance.

To demonstrate the effectiveness of jointly optimizing the IMU model and MPC, we investigate four scenarios: (1) IMU+MPC: classic IMU integrator with a regular d-MPC; (2) IMU$^+$+MPC: data-driven IMU model with a regular d-MPC; (3) IMU+MPC$^+$: classic IMU integrator with a d-MPC with learnable parameters, e.g. mass, moment of inertia (MOI); and (4) iMPC (ours): IMU$^+$ with MPC$^+$ trained with IL framework, where ``IMU$^+$'' refers to the data-driven IMU denoising and integration model from our previous work \citep{qiu2023airimu}. Additionally, for the wind disturbance test, we also implement a RL approach with a commonly used Proximal Policy Optimization (PPO) \citep{schulman2017proximal} as the baseline. Specifically, we take pre-integrated IMU measurements as the observation, output UAV action (thrust and torques) in OpenAI gym \citep{brockman2016openai}, and designed the reward to minimize the difference between the current and target UAV pose.

We use three metrics to evaluate the control performance: settling time (\textit{ST}), root-mean-square error (\textit{RMSE}), and steady-state error (\textit{SSE}).
Specifically, \textit{ST} is the time for the UAV to enter and remain within $\pm 1.5 \degree$ of its final steady attitude, measuring how quickly the system settles; \textit{RMSE} is the root-mean-square of the difference between estimated and the desired attitude; and \textit{SSE} is the absolute difference between the steady and desired attitude, evaluating the control accuracy. All experiments are repeated ten times for generalizability. 

\subsection{Results} \label{subsec:results}

\begin{table}[t]
    \centering
    \caption{The performance of UAV attitude control under different initial conditions. The ``IMU'' means the RMSE (unit: $\times 10^{-3}\mathrm{rad}$) of attitude estimation for the corresponding method.}
    \label{tab:impc:init}
    \resizebox{0.98\linewidth}{!}{
    \begin{tabular}{c|ccc|ccc|ccc|ccc}
        \toprule
         & \multicolumn{3}{|c|}{IMU+MPC} & \multicolumn{3}{|c|}{IMU$^+$+MPC} & \multicolumn{3}{|c|}{IMU+MPC$^+$} & \multicolumn{3}{|c}{\textbf{iMPC (ours)}} \\
        \midrule
        Initial & 10\degree & 15\degree & 20\degree & 10\degree & 15\degree & 20\degree & 10\degree & 15\degree & 20\degree & 10\degree & 15\degree & 20\degree \\
        \midrule
        \textit{ST} ($\second$) & $0.287$ & $0.330$ & $0.336$ & $0.281$ & $0.299$ & $0.318$ & $0.283$ & $0.321$ & $0.324$ & $\mathbf{0.275}$ & $\mathbf{0.296}$ & $\mathbf{0.317}$ \\
        \textit{RMSE} (\degree) & $0.745$ & $0.726$ & $0.728$ & $0.692$ & $0.691$ & $0.685$ & $0.726$ & $0.721$ & $0.715$ & $\mathbf{0.691}$ & $\mathbf{0.688}$ & $\mathbf{0.684}$ \\
        \textit{SSE} (\degree) & $0.250$ & $0.262$ & $0.263$ & $0.193$ & $0.213$ & $0.217$ & $0.216$ & $0.230$ & $0.224$ & $\mathbf{0.185}$ & $\mathbf{0.201}$ & $\mathbf{0.208}$ \\
        IMU & $7.730$ & $8.390$ & $8.370$ & $6.470$ & $7.020$ & $7.270$ & $7.370$ & $7.980$ & $8.090$ & $\mathbf{6.310}$ & $\mathbf{6.800}$ & $\mathbf{7.010}$\\
        \bottomrule
    \end{tabular}}
\vspace{-4mm}
\end{table}

\begin{figure}[t]
    \centering
    \includegraphics[width=0.98\linewidth]{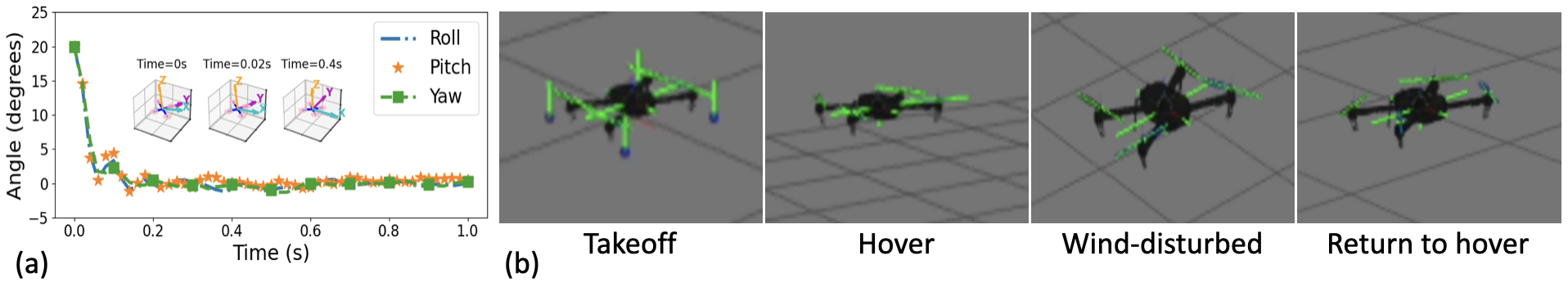}
    \caption{UAV Performances. (a) The UAV attitude quickly returns to a stable hover for an initial condition of 20\degree~using iMPC. (b) Snapshots of iMPC under 20 m/s wind disturbance in Gazebo, including takeoff, hover, being disturbed by the wind, and returning to the hover.}
    \label{fig:mpc-response}
\vspace{-6mm}
\end{figure} 

\textbf{\textit{Different Initial Conditions}}: 
We first evaluate system control performance under various initial conditions using our customized simulation, which offers flexibility and avoids the physical setup challenges of Gazebo. \cref{tab:impc:init} compares performance across scenarios with initial attitudes ranging from 10\degree~to 20\degree~for all three Euler angles (roll, pitch, and yaw). Our method reliably stabilizes the UAV under all tested conditions, with all exhibiting stable performance and negligible standard deviations. Compared to baseline methods, our framework achieves lower \textit{SSE}, \textit{RMSE}, and \textit{ST}. \cref{fig:mpc-response} demonstrates how rapidly the UAV returns to a stable hover from an initial attitude of 20\degree~using our approach. To investigate how the LL MPC assists UL IMU parameter learning, we include IMU attitude estimation results in the last column of \cref{tab:impc:init}. Thanks to the physics knowledge from dynamics as well as the LL MPC, the denoising and prediction performance of the IMU module get significantly improved compared to the separately training. Consequently, our self-supervised learning framework improves both IMU network learning and overall control performance.

\begin{table}[t]
    \centering
    \caption{UAV attitude control performance under different \textbf{impulse} and \textbf{step wind} disturbances.}
    \label{tab:combined-wind}
    \resizebox{0.98\linewidth}{!}{\begin{tabular}{c|c|c|cccc|cccc}
        \toprule
         & & & \multicolumn{4}{|c|}{\textbf{Impulse Wind Disturbance}} & \multicolumn{4}{c}{\textbf{Step Wind Disturbance}} \\
        \midrule
         & Wind & Methods & \textit{ST} ($\second$) & \textit{RMSE} (\degree) & \textit{SSE} (\degree) & IMU & \textit{ST} ($\second$) & \textit{RMSE} (\degree) & \textit{SSE} (\degree) & IMU \\
        \midrule
        \multirow{12}{*}{\centering Customized}
        & \multirow{4}{*}{\centering 10 \meter/\second}
        & IMU+MPC & $0.406$ & $0.360$ & $0.169$ & $9.190$ & $0.596$ & $0.396$ & $0.189$ & $6.870$ \\ 
        & & IMU$^+$+MPC & $0.392$ & $0.355$ & $0.142$ & $9.010$ & $0.594$ & $0.373$ & $0.183$ & $6.110$ \\ 
        & & IMU+MPC$^+$ & $0.406$ & $0.360$ & $0.168$ & $9.180$ & $0.588$ & $0.392$ & $0.179$ & $6.830$ \\ 
        & & RL (PPO) & $0.563$ & $1.721$ & $1.398$ & \ding{55} & $0.747$ & $1.785$ & $1.454$ & \ding{55} \\
        & & \textbf{iMPC (ours)} & $\mathbf{0.368}$ & $\mathbf{0.353}$ & $\mathbf{0.132}$ & $\mathbf{8.660}$ & $\mathbf{0.574}$ & $\mathbf{0.352}$ & $\mathbf{0.168}$ & $\mathbf{5.990}$ \\ 
        \cmidrule{2-11}
        & \multirow{4}{*}{\centering 15 \meter/\second} 
        & IMU+MPC & $0.454$ & $0.328$ & $0.120$ & $8.160$ & $0.684$ & $0.337$ & $0.172$ & $7.190$ \\ 
        & & IMU$^+$+MPC & $0.450$ & $0.316$ & $0.065$ & $7.730$ & $0.670$ & $0.309$ & $0.167$ & $6.740$ \\ 
        & & IMU+MPC$^+$ & $0.454$ & $0.325$ & $0.117$ & $8.140$ & $0.684$ & $0.337$ & $0.171$ & $7.180$ \\ 
        & & RL (PPO) & \ding{55} & \ding{55} & \ding{55} & \ding{55} & \ding{55} & \ding{55} & \ding{55} & \ding{55} \\
        & & \textbf{iMPC (ours)} & $\mathbf{0.430}$ & $\mathbf{0.312}$ & $\mathbf{0.060}$ & $\mathbf{7.570}$ & $\mathbf{0.620}$ & $\mathbf{0.297}$ & $\mathbf{0.149}$ & $\mathbf{6.730}$ \\ 
        \cmidrule{2-11}
        & \multirow{4}{*}{\centering 20 \meter/\second} 
        & IMU+MPC & $0.486$ & $0.361$ & $0.186$ & $6.550$ & $0.704$ & $0.376$ & $0.240$ & $6.300$ \\ 
        & & IMU$^+$+MPC & $0.476$ & $0.356$ & $0.139$ & $6.330$ & $0.690$ & $0.317$ & $0.173$ & $6.120$ \\ 
        & & IMU+MPC$^+$ & $0.484$ & $0.361$ & $0.185$ & $6.520$ & $0.704$ & $0.372$ & $0.253$ & $6.300$ \\ 
        & & RL (PPO) & \ding{55} & \ding{55} & \ding{55} & \ding{55} & \ding{55} & \ding{55} & \ding{55} & \ding{55} \\
        & & \textbf{iMPC (ours)} & $\mathbf{0.470}$ & $\mathbf{0.354}$ & $\mathbf{0.135}$ & $\mathbf{6.160}$ & $\mathbf{0.676}$ & $\mathbf{0.311}$ & $\mathbf{0.150}$ & $\mathbf{6.110}$ \\ 
        \midrule
        \multirow{12}{*}{\centering Gazebo}
        & \multirow{4}{*}{\centering 10 \meter/\second} 
        & IMU+MPC & $0.654$ & $0.644$ & $0.259$ & $10.180$ & $0.778$ & $0.665$ & $0.311$ & $10.540$ \\ 
        & & IMU$^+$+MPC & $0.622$ & $0.636$ & $0.246$ & $9.970$ & $0.766$ & $0.646$ & $0.302$ & $10.330$ \\ 
        & & IMU+MPC$^+$ & $0.649$ & $0.342$ & $0.255$ & $10.140$ & $0.769$ & $0.665$ & $0.307$ & $10.530$ \\ 
        & & \textbf{iMPC (ours)} & $\mathbf{0.616}$ & $\mathbf{0.630}$ & $\mathbf{0.238}$ & $\mathbf{9.560}$ & $\mathbf{0.759}$ & $\mathbf{0.640}$ & $\mathbf{0.289}$ & $\mathbf{10.120}$ \\ 
        \cmidrule{2-11}
        & \multirow{4}{*}{\centering 15 \meter/\second} 
        & IMU+MPC & $0.688$ & $0.658$ & $0.240$ & $8.880$ & $0.843$ & $0.673$ & $0.324$ & $9.010$ \\ 
        & & IMU$^+$+MPC & $0.647$ & $0.641$ & $0.211$ & $8.270$ & $0.841$ & $0.638$ & $0.315$ & $8.950$ \\ 
        & & IMU+MPC$^+$ & $0.668$ & $0.652$ & $0.237$ & $8.800$ & $0.835$ & $0.671$ & $0.311$ & $9.010$ \\ 
        & & \textbf{iMPC (ours)} & $\mathbf{0.635}$ & $\mathbf{0.637}$ & $\mathbf{0.206}$ & $\mathbf{8.150}$ & $\mathbf{0.827}$ & $\mathbf{0.626}$ & $\mathbf{0.306}$ & $\mathbf{8.740}$ \\ 
        \cmidrule{2-11}
        & \multirow{4}{*}{\centering 20 \meter/\second} 
        & IMU+MPC & $0.729$ & $0.667$ & $0.275$ & $7.490$ & $0.996$ & $0.690$ & $0.322$ & $7.090$ \\ 
        & & IMU$^+$+MPC & $0.718$ & $0.665$ & $0.254$ & $7.230$ & $0.975$ & $0.655$ & $0.298$ & $6.800$ \\ 
        & & IMU+MPC$^+$ & $0.720$ & $0.667$ & $0.273$ & $7.440$ & $0.996$ & $0.684$ & $0.320$ & $7.090$ \\ 
        & & \textbf{iMPC (ours)} & $\mathbf{0.711}$ & $\mathbf{0.659}$ & $\mathbf{0.243}$ & $\mathbf{7.210}$ & $\mathbf{0.963}$ & $\mathbf{0.647}$ & $\mathbf{0.293}$ & $\mathbf{6.550}$ \\
        \bottomrule
    \end{tabular}}
\end{table}

\begin{figure}[t]
  \centering
  \subfigure[UAV pitch angle response when encountering an \textbf{impulse wind} disturbance at 0.2~s for different speeds. PPO loses control at a wind speed of 15 \meter/\second, while iMPC loses control at 45 \meter/\second. \label{fig:wind-impulse}]{
    \includegraphics[width=0.48\linewidth]{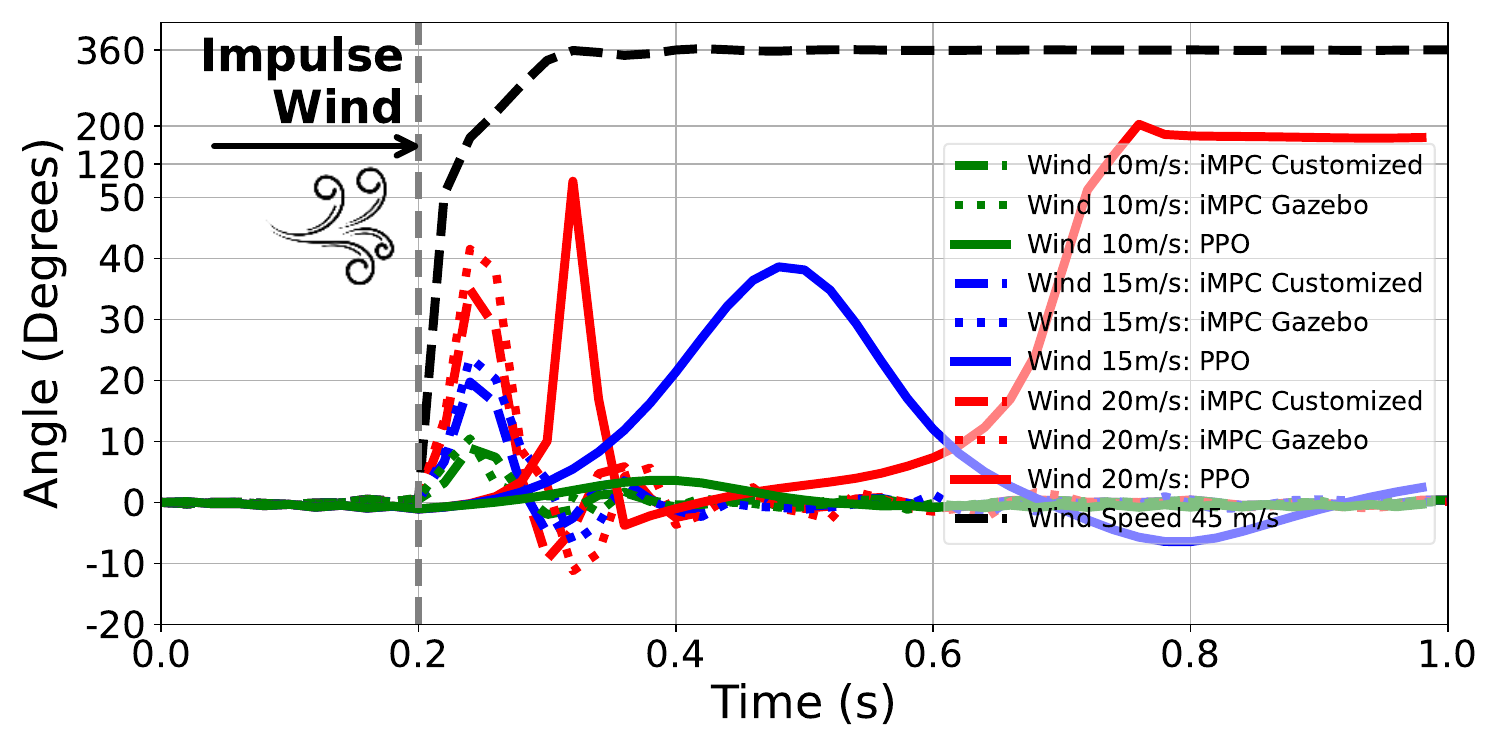}
  } \hfill
  \subfigure[UAV pitch angle response when the encountering a \textbf{step wind} disturbance at 0.2~s and lasting for 0.3~s for different speeds. PPO loses control at a wind speed of 15 \meter/\second, while iMPC loses control at 35 \meter/\second. \label{fig:wind-step}]{
    \includegraphics[width=0.48\linewidth]{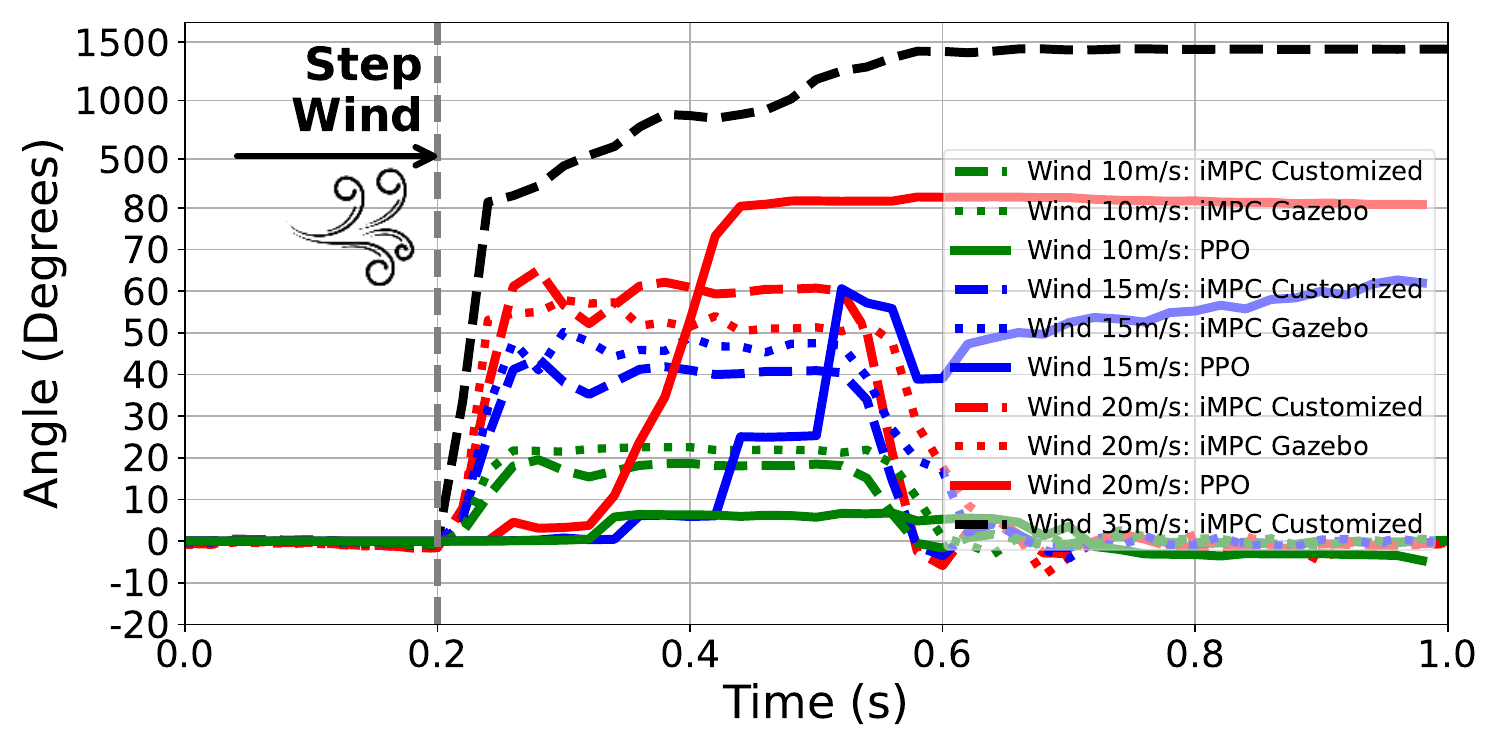}
  }
  \caption{Control performance of iMPC and RL (PPO) under different levels of wind disturbance.}
  \label{fig:wind_gust}
  \vspace{-5mm}
\end{figure}

\textbf{\textit{Wind Disturbance}}: 
To validate robustness, we evaluate system performance under wind disturbances in both customized and Gazebo simulations. We conducted two experiments: (1) impulse wind gusts opposite to the UAV's heading during hover at wind speeds of $10/15/20/45\ \meter/\second$, and (2) step winds in the same direction during hover for 0.3~\second~at speeds of $10/15/20/35\ \meter/\second$, beginning 0.2~\second~after stable hover. During RL (PPO) training, we apply external wind disturbances of up to 10m/s. In contrast, our hybrid MPC approach provides robustness without exposure to diverse wind conditions during training.  \cref{tab:combined-wind} compares iMPC with other methods, showing that iMPC yields lower \textit{SSE}, shorter \textit{ST}, and more accurate attitude prediction. Particularly compared to IMU+MPC and IMU$^+$+MPC, our method significantly improves IMU denoising and reduces RMSE. Even under substantial disturbances, iMPC ensures robust attitude control, demonstrating its ability to simultaneously enhance both MPC parameter learning and IMU denoising and prediction.

\cref{fig:mpc-response}~(b) shows the entire process of the UAV initially hovering, being disturbed by wind, and returning to stable hover, demonstrating that the UAV in Gazebo simulation can still resist wind and recover quickly, even under $20\ \meter/\second$ wind.
UAV pitch angle responses using our iMPC and PPO under different wind disturbances are shown in \cref{fig:wind_gust}.
The wind affects the UAV attitude with different peak values under different speeds. Even under $20\ \meter/\second$ impulse or step wind, the UAV overcomes the disturbance and quickly returns to hover. Note that the system tolerates higher speeds under impulse wind compared to step wind, due to the latter's longer duration. When the wind speed is too high, actuator limits prevent compensation, leading to failure and crash.
Compared to PPO, iMPC exhibits notably better performance, especially under wind disturbances not seen during training. While PPO performs comparably under seen wind conditions, it completely loses control under out-of-distribution disturbances. These results highlight the robustness of our framework, benefiting from a hybrid structure that integrates prior physical knowledge.

\begin{table}[t]
    \centering
    \caption{The learned UAV MOI and mass error using our method (iMPC).}
    \label{tab:learned_params}
    \resizebox{0.8\linewidth}{!}{
    \begin{tabular}{c|c|cc|cc|cc|cc}
    \toprule
    & Initial Offset & Initial & Error & Initial & Error & Initial & Error & Initial & Error \\ 
    \midrule
     MOI & 50\% & $ 0 \degree $ & $ 0.96 \%$ & $ 10 \degree $ & $ 2.67 \%$ & $ 15 \degree $ & $ 3.41 \%$ & $ 20 \degree $ & $ 2.22 \%$    \\
     Mass & 50\% & $ 0 \degree $ & $ 1.69\%$ & $ 10 \degree $ & $ 0.85\%$ & $ 15 \degree $ & $ 1.43\%$ & $ 20 \degree $ & $ 0.32\%$    \\
    \bottomrule
    \end{tabular}}
\vspace{-5mm}
\end{table}

\textbf{\textit{Learned Dynamics Parameters}}:
We also evaluate the learning performance in the d-MPC. In particular, we treat the UAV mass and moment of inertia (MOI) as the learnable parameters.
The goal is to validate the final estimated mass and MOI match the real value after the learning process. To illustrate this, we set an initial guess of the vehicle MOI and mass with 50\% offset from its true value and show the final estimated error using our method in \cref{tab:learned_params}.
It can be seen that iMPC achieves a final MOI error of less than 3.5\%, and a final mass error of less than 1.7\%. 
By jointly considering the performance in \cref{tab:impc:init} and \cref{tab:learned_params} and compare iMPC with IMU$^+$+MPC, and IMU+MPC$^+$ with IMU+MPC, where ``MPC$^+$'' indicates the learned MOI and mass are in the control loop, we conclude that better-learned dynamics parameters lead to better attitude control performance, with smaller SSE and settling time.

\textbf{\textit{Efficiency Analysis}}:
Imperative MPC matches IMU + MPC in runtime because MPC dominates computation and the learning-based IMU inference adds negligible overhead. It runs at 50 Hz with a 200 Hz IMU , which is at the same level as standard MPC. Compared to RL, although the inference time of iMPC is longer as expected due to the need of solving an online optimization, its physics-based prior helps overcome the typical low sampling efficiency issue of RL during training. In fact, our experiment shows the PPO needs 221.28\% more training time.

\section{Conclusions} \label{sec:con}

This paper proposes a novel end-to-end self-supervised learning framework that integrates a learning-based IO module with d-MPC for UAV attitude control. The framework addresses the challenges of nonlinear dynamics and unpredictable external disturbances by combining the robustness of data-driven perception with the interpretability of physics-based control. 
The problem was formulated as a bi-level optimization, where the inner MPC optimizes control actions and the upper level minimizes discrepancy between real-world and the predicted performance. The framework is thus end-to-end and can be trained in a self-supervised manner. 
The framework was validated in both a customized Python and PX4 Gazebo simulations. Results show the effectiveness even under strong external wind of up to 20 m/s, achieving a steady-state error with an accuracy of 0.243 degrees. The joint learning mechanism simultaneously enhances both the MPC parameter learning and IMU denoising and prediction performance. Future works includes directly deploying the PX4-based integration in to the real world UAVs and considering uncertainty in the perception module.

\acks{We thank Dr. Chen Wang and Qihang Li at the University at Buffalo for their suggestion and support to this research.}

\end{document}